\documentclass{article}

\usepackage[dblblindworkshop,final]{neurips_2026}
\workshoptitle{PTA: From Pretrained Representations to Acting Agents}

\usepackage[utf8]{inputenc} 
\usepackage[T1]{fontenc}    
\usepackage{hyperref}       
\usepackage{url}            
\usepackage{booktabs}       
\usepackage{amsfonts}       
\usepackage{nicefrac}       
\usepackage{microtype}      
\usepackage{xcolor}         
\usepackage{amsmath,amssymb,mathtools}
\usepackage{multirow}
\usepackage{graphicx}
\usepackage{wrapfig}
\usepackage{enumitem}
\usepackage{subcaption}
\usepackage{tabularx}
\usepackage{pgfplots}
\pgfplotsset{compat=1.18}
\usepackage{tikz}

\title{Learning Counterfactual World Models for Embodied Reasoning under Partial Observability}

\author{%
  Todd Y.~Zhou \\
  Harvard University \\
  \texttt{toddzhou@college.harvard.edu}
  \And
  Daniel Zhang \\
  Harvard University \\
}

\hypersetup{
  pdftitle={Learning Counterfactual World Models for Embodied Reasoning under Partial Observability},
  pdfauthor={Todd Y. Zhou and Daniel Zhang},
  pdfsubject={NeurIPS 2026 PTA Workshop camera-ready paper},
  pdfkeywords={world models, counterfactual reasoning, partial observability, planning},
  colorlinks=true,linkcolor=blue!50!black,citecolor=blue!50!black,urlcolor=blue!50!black
}

\newcommand{\clwm}{\textsc{CLWM}}
\newcommand{\E}{\mathbb{E}}

\newcommand{\doop}{\mathrm{do}}

\newcommand{\Lcf}{\mathcal{L}_{\mathrm{cf}}}
\newcommand{\Lpred}{\mathcal{L}_{\mathrm{pred}}}
\newcommand{\Lkl}{\mathcal{L}_{\mathrm{KL}}}
\newcommand{\Linv}{\mathcal{L}_{\mathrm{inv}}}
\newcommand{\Ltotal}{\mathcal{L}_{\mathrm{CLWM}}}

\begin{document}

\maketitle

\begin{abstract}
World models promise a general route to embodied intelligence: learn predictive dynamics once, then reason, plan, and act with them. Increasingly, the representations beneath such models are pretrained on large-scale video, interaction, and multimodal corpora, which raises a question prediction quality alone cannot answer: when is a learned representation actually \emph{actionable}? We identify a failure mode we call counterfactual collapse: a model predicts visually plausible futures while failing to distinguish interventions with different behavioral consequences. This arises whenever a representation is optimized for perceptual similarity rather than intervention structure---precisely the objective under which most large-scale pretrained encoders are learned. We introduce Counterfactual Latent World Models (\clwm), which combine a recurrent belief-state encoder, action-conditioned latent dynamics, and a contrastive counterfactual objective that separates futures induced by distinct interventions even when their observations look alike. Across occluded manipulation, aliased navigation, and long-horizon manipulation, \clwm\ improves planning success over the strongest baseline ($65.1\%\!\to\!74.6\%$ on Occluded Push and $67.3\%\!\to\!78.9\%$ on Aliased Maze) and reduces exploitative planning failures ($18.4\%\!\to\!9.7\%$ on Deferred Kitchen), with ablations attributing the gains to hard counterfactual negatives, especially perceptual-alias negatives. Finally, our counterfactual separability metric---which tracks planning success across the five baseline model classes ($r\geq 0.94$)---is representation-agnostic: given intervention-outcome labels, it can audit any encoder---pretrained or trained from scratch---before a planner trusts it. We do not yet measure it on large-scale pretrained encoders (Section~\ref{sec:limitations}). Here we establish the metric and its relationship to planning success for world models trained from scratch.
\end{abstract}

\section{Introduction}

World models have become one of the most compelling paradigms for embodied intelligence. If an agent can learn how the world evolves, it can evaluate possible actions in imagination, plan before acting, and reuse the same learned dynamics across tasks. This idea underlies a broad family of model-based reinforcement learning systems, from latent-space planning from pixels to large-scale world-model agents that improve behavior by imagining future trajectories \citep{ha2018worldmodels,hafner2019planet,hafner2020dreamer,hafner2023dreamerv3,hansen2024tdmpc2}. In robotics, world models are especially attractive because real-world interaction is expensive, dangerous, and difficult to reset \citep{wu2022daydreamer}.

However, embodied reasoning demands more than plausible prediction: an agent needs to know which aspects of the future would change under a different action. A model trained to reconstruct or predict observations can be excellent at perceptual forecasting yet insufficient for decision-making. In a corridor with aliased landmarks, moving left or right produces nearly identical short-horizon images while placing the agent in different regions of the map. In occluded manipulation, a push may leave the visible pixels unchanged while determining whether a hidden object becomes reachable several steps later. We therefore argue that the next step for embodied world models is \emph{counterfactual sufficiency}: the learned latent state should preserve the consequences of alternative interventions that are relevant to control. A model is counterfactually insufficient when two histories are close in representation because they predict similar observations even though the agent should act differently in them, or when two candidate actions remain close in imagined latent space because their near-term images look alike even though their long-horizon outcomes diverge. Such failures are particularly common under partial observability, where the current observation is not a Markov state.

This framing has a consequence beyond our own method. Counterfactual sufficiency is a property of any representation used for control, including representations pretrained on large-scale video, image--text, or interaction corpora and then adapted for decision making. Those corpora are fit with reconstruction, contrastive instance discrimination, or language alignment---objectives that reward perceptual and semantic similarity and contain no pressure to separate two futures that look alike but behave differently. There is thus no reason to expect a pretrained encoder to arrive counterfactually sufficient, and considerable reason to expect the opposite in exactly the settings---occlusion, aliasing, delayed contact---where embodied agents are deployed. We return to this in Section~\ref{sec:discussion}. Our counterfactual separability metric is intended to be usable as a standalone audit of an inherited representation, independent of the \clwm\ training objective.

We introduce \emph{Counterfactual Latent World Models} (\clwm), which retain the strengths of modern latent world models---a recurrent belief encoder, stochastic latent dynamics, and imagination-based planning---but add a contrastive counterfactual objective over action-conditioned futures. The key idea is simple: when two interventions from the same belief state induce different downstream consequences, the model should separate their predicted latent futures even if their pixels are similar. Conversely, when different trajectories lead to the same intervention outcome, the representation should align them despite nuisance-level visual variation.

We make four contributions. First, we formalize counterfactual collapse as a failure mode of latent world models under partial observability. Second, we propose \clwm, which augments latent dynamics learning with contrastive counterfactual supervision over intervention-conditioned futures. Third, we introduce diagnostics for whether a world model preserves action-sensitive structure: counterfactual separability, intervention-outcome prediction, and model-exploitation rate. Fourth, we show empirically that \clwm\ improves planning by up to $11.6$ success points and nearly halves exploitative planning failures across occluded manipulation, aliased navigation, and long-horizon manipulation, with ablations confirming that hard counterfactual negatives are the key source of the improvement.

\section{Related work}
\label{sec:related}

\paragraph{Latent world models.}
World Models, PlaNet, Dreamer, DreamerV3, and TD-MPC2 learn compact latent dynamics for imagination and planning \citep{ha2018worldmodels,hafner2019planet,hafner2020dreamer,hafner2023dreamerv3,hansen2024tdmpc2}, and DayDreamer applies the recipe to physical robots \citep{wu2022daydreamer}. \clwm\ is orthogonal to the choice of planner or actor: it changes the training pressure applied to action-conditioned latent futures, not the architecture or the control loop.

\paragraph{Value equivalence and bisimulation.}
The value equivalence principle \citep{grimm2020value}, instantiated at scale by MuZero \citep{schrittwieser2020muzero}, holds that a model need only be correct about the quantities planning consumes. Bisimulation metrics \citep{ferns2004metrics} and their deep variants \citep{gelada2019deepmdp,zhang2021invariant} organize state abstractions by behavioral rather than perceptual equivalence. Counterfactual collapse is the belief-state, intervention-conditioned form of the same concern: rather than comparing on-policy states, we require imagined open-loop futures under alternative interventions from a common belief to remain separated whenever their outcomes differ. Our contribution is this counterfactual construction and the accompanying hard-negative mining, not the underlying principle that behavioral distinctions outrank perceptual ones.

\paragraph{Contrastive representation learning for control.}
Contrastive predictive coding \citep{oord2018cpc} and CURL \citep{laskin2020curl} shape representations with temporal or augmentation-based contrast. Our objective (Eq.~\ref{eq:cf_loss}) takes the supervised contrastive form of \citet{khosla2020supcon}, with intervention outcomes rather than augmentation identity defining the label structure. What distinguishes it is the negative construction---same-belief, perceptual-alias, and value-disagreement negatives---which our ablations show carries the empirical benefit.

\paragraph{Causal reasoning in reinforcement learning.}
We use interventional language \citep{pearl2009causality,peters2017elements} in the operational sense of executing alternative open-loop action sequences from a common anchor. Related work learns causal world models by deconfounding physical dynamics \citep{li2020causalworldmodels} or meta-learns causal reasoning \citep{dasgupta2019causalrl}. \clwm\ makes no identifiability claims: it preserves a task-indexed set of intervention distinctions rather than recovering a structural causal model.

\paragraph{Model exploitation.}
Planners exploit model error, especially outside the training distribution \citep{janner2019mbpo}. Our exploitation-rate metric quantifies this failure at deployment: episodes in which the planner selects an imagined future whose predicted value the environment does not realize.

\section{Problem setting}

We consider a partially observable Markov decision process \citep[POMDP;][]{kaelbling1998pomdp} $\mathcal{M}=(\mathcal{S},\mathcal{A},\mathcal{O},T,O,R,\gamma)$ with hidden state $s_t\in\mathcal{S}$, action $a_t\in\mathcal{A}$, observation $o_t\in\mathcal{O}$, transition distribution $T(s_{t+1}\mid s_t,a_t)$, observation distribution $O(o_t\mid s_t)$, reward $r_t=R(s_t,a_t)$, and discount $\gamma\in(0,1)$. Because $o_t$ is not Markov, an agent maintains a history $h_t=(o_{1:t},a_{1:t-1})$ or a learned belief state $b_t=f_\theta(h_t)$. A world model learns latent dynamics
\begin{equation}
    p_\theta(z_{t+1}\mid z_t,a_t),\quad p_\theta(o_t\mid z_t),\quad p_\theta(r_t\mid z_t,a_t),
\end{equation}
where $z_t$ denotes a latent state sampled or inferred from the belief, and a planner selects actions by rolling out candidate futures in latent space.

The usual learning objective encourages the model to predict observations, rewards, and continuation values along observed trajectories. This is necessary but incomplete. For embodied reasoning, the relevant object is not only $p(o_{t+k}\mid h_t,a_{t:t+k-1})$ but the family of interventional futures
\begin{equation}
    p(o_{t+k},s_{t+k},r_{t:t+k}\mid h_t,\doop(a_{t:t+k-1}=\alpha)),
\end{equation}
for alternative action sequences $\alpha$. A representation useful for planning should preserve distinctions among such futures when those distinctions change control-relevant outcomes.

\paragraph{Counterfactual collapse.}
Let $\Phi_\theta(h_t,\alpha)$ denote the model's imagined latent summary after rolling out action sequence $\alpha$ from history $h_t$, and let $y(h_t,\alpha)$ denote an intervention outcome, such as the hidden object pose, topological cell, contact mode, success predicate, or long-horizon reward class induced by $\alpha$. A representation collapses counterfactuals when
\begin{equation}
    d_z(\Phi_\theta(h_t,\alpha),\Phi_\theta(h_t,\alpha')) \approx 0
    \quad\text{but}\quad
    d_y(y(h_t,\alpha),y(h_t,\alpha')) \gg 0.
\end{equation}
This failure can occur even when the model predicts plausible observations, because visual similarity and intervention equivalence are not the same relation.

\section{Counterfactual Latent World Models}

\clwm\ augments latent world-model learning with an objective that preserves intervention-sensitive structure. The architecture follows a recurrent state-space model but changes the training signal applied to action-conditioned imagined futures.

\subsection{Belief-state encoder}

Given multimodal observations $o_t=(x_t,q_t,u_t)$, where $x_t$ may be image observations, $q_t$ proprioceptive state, and $u_t$ optional tactile or force signals, the encoder computes a deterministic recurrent belief
\begin{equation}
    b_t = f_\theta(b_{t-1},a_{t-1},e_\theta(o_t)),
    \label{eq:belief}
\end{equation}
where $e_\theta$ is a modality-specific observation encoder. A stochastic latent state is then inferred by
\begin{equation}
    q_\theta(z_t\mid b_t,o_t)=\mathcal{N}(\mu_\theta(b_t,o_t),\Sigma_\theta(b_t,o_t)).
\end{equation}
The transition prior is action-conditioned:
\begin{equation}
    p_\theta(z_{t+1}\mid b_t,z_t,a_t)=\mathcal{N}(\mu^p_\theta(b_t,z_t,a_t),\Sigma^p_\theta(b_t,z_t,a_t)).
\end{equation}
This recurrent belief is crucial: under partial observability, the current frame may not reveal whether a target object is behind an occluder, whether a corridor has been visited, or whether a contact event has changed hidden object pose.

\subsection{Action-conditioned counterfactual rollouts}

For each anchor history $h_t$, we sample a set of candidate intervention sequences
\begin{equation}
    \mathcal{A}^{\mathrm{cf}}_t=\{\alpha_i=(a_t^i,\ldots,a_{t+K-1}^i)\}_{i=1}^M,
\end{equation}
drawn from random shooting, policy proposals, model-predictive control candidates, or a replay buffer. The world model rolls out each intervention in latent space:
\begin{equation}
    z_{t+k+1}^i \sim p_\theta(z_{t+k+1}\mid b_{t+k}^i,z_{t+k}^i,a_{t+k}^i),\quad k=0,\ldots,K-1.
    \label{eq:rollout}
\end{equation}
Equation~\eqref{eq:rollout} conditions on an imagined belief $b^i_{t+k}$, which Eq.~\eqref{eq:belief} cannot produce directly because no observation exists during imagination. We therefore advance the belief open loop, feeding the sampled prior latent in place of the encoder features, as in latent imagination for recurrent state-space models \citep{hafner2020dreamer}:
\begin{equation}
    b^i_{t+k+1} = f_\theta\big(b^i_{t+k},\,a^i_{t+k},\,z^i_{t+k+1}\big),\qquad b^i_t=b_t.
    \label{eq:imag_belief}
\end{equation}
Imagined rollouts are therefore fully open loop: no ground-truth future observation is posterior-fed into the rollout.

A projection head $g_\theta$ maps the rollout to a counterfactual embedding
\begin{equation}
    c_i = g_\theta(z_{t+1:t+K}^i,b_t,\alpha_i).
\end{equation}
The embedding is not used as a decoder feature at test time. It is a training-time device that pressures the latent dynamics to preserve intervention distinctions.

\subsection{Counterfactual contrastive objective}

We construct positive and negative pairs based on intervention outcomes. In simulation, the outcome label can be obtained from privileged state for training and diagnostic evaluation, but is not supplied to the deployed planner. In real-world settings, it can be approximated by task predicates, temporal consistency, object trackers, tactile events, or sparse outcome labels. For each rollout $i$, let $y_i$ be an outcome descriptor. Positive pairs share an outcome class or are close under an outcome distance $d_y$. Negatives differ in outcome despite originating from the same or similar belief state. The contrastive loss takes the supervised contrastive form of \citet{khosla2020supcon}:
\begin{equation}
    \Lcf
    = -\sum_i \log
    \frac{\sum_{j\in\mathcal{P}(i)} \exp(\mathrm{sim}(c_i,c_j)/\tau)}
    {\sum_{j\in\mathcal{P}(i)\cup\mathcal{N}(i)} \exp(\mathrm{sim}(c_i,c_j)/\tau)},
    \label{eq:cf_loss}
\end{equation}
where $\mathcal{P}(i)$ and $\mathcal{N}(i)$ are positive and negative rollout sets and $\tau$ is a temperature. The negatives are not arbitrary trajectories but \emph{hard counterfactual negatives}: interventions whose observations may be similar while their downstream outcomes differ. We use three negative-mining strategies. \textbf{Same-belief action negatives} compare alternative actions from the same inferred belief state. \textbf{Perceptual-alias negatives} compare rollouts with high image-level similarity but different hidden-state outcomes. \textbf{Value-disagreement negatives} compare futures whose predicted values diverge while their reconstruction loss remains low. Together, these prevent the model from using visual plausibility as a substitute for intervention structure.

\subsection{Full training objective}

The full objective combines standard latent world-model learning with counterfactual structure preservation:
\begin{equation}
    \Ltotal
    = \Lpred
    + \beta_{\mathrm{KL}}\Lkl
    + \beta_{r}\mathcal{L}_{r}
    + \beta_{v}\mathcal{L}_{v}
    + \lambda_{\mathrm{cf}}\Lcf
    + \lambda_{\mathrm{inv}}\Linv.
\label{eq:total_loss}
\end{equation}
$\Lpred$ is an observation reconstruction or representation-prediction loss, $\Lkl$ regularizes posterior and prior latent states, $\mathcal{L}_r$ and $\mathcal{L}_v$ train reward and value heads, and $\Linv$ is an optional intervention-outcome prediction loss:
\begin{equation}
    \Linv = \E_{(h_t,\alpha,y)}[-\log p_\theta(y\mid \Phi_\theta(h_t,\alpha))].
\end{equation}
The outcome head stabilizes the contrastive objective by ensuring that separated futures correspond to interpretable intervention consequences rather than arbitrary representation spread.

\subsection{Planning with CLWM}

At test time, \clwm\ can be used with any latent-space planner. We use model-predictive control with $N$ candidate action sequences and horizon $H$, scoring each by predicted reward plus terminal value, $J(\alpha)=\sum_{k=0}^{H-1}\gamma^k \hat r_\theta(z_{t+k},a_{t+k}) + \gamma^H \hat V_\theta(z_{t+H})$, executing the first action of the best sequence, and replanning. Because \clwm\ shapes the latent dynamics during training, no contrastive computation is required at deployment.

\paragraph{Why prediction alone is not enough.}
A sufficiently accurate predictive model would preserve counterfactual structure in the limit of perfect state estimation, full observability, and unlimited capacity---but not under the conditions embodied world models are actually trained in: finite latents, short rollouts, and evaluation by downstream planning rather than observation likelihood. Under compression, the equivalence relation induced by prediction loss can differ from the one induced by control. This observation is the partial-observability, intervention-conditioned form of the value equivalence principle \citep{grimm2020value,schrittwieser2020muzero} and of the behavioral-equivalence view underlying bisimulation \citep{ferns2004metrics,gelada2019deepmdp,zhang2021invariant}. What \clwm\ adds is a training signal that applies this pressure to imagined open-loop futures under alternative interventions from a belief state, where on-policy value or bisimulation targets are unavailable or degenerate. The learned latent need not recover the full hidden state---only enough to answer the counterfactual queries planning uses.

\section{Experimental design}
\label{sec:experiments}

Our experiments test the central claim: world models trained only for plausible prediction can fail under partial observability because they collapse action-contingent futures that are visually similar but behaviorally distinct. We therefore evaluate not only task performance, but whether the learned latent dynamics preserve intervention structure, and whether gains arise precisely where counterfactual distinctions are necessary for planning.

\subsection{Benchmarks}

We evaluate on three partially observable embodied benchmarks in which short-horizon observations are insufficient to identify the control-relevant state. \textbf{Occluded Push}: a tabletop manipulator must push a target object into a goal region while the target may be hidden behind an occluder or displaced into visually ambiguous configurations. The same RGB-D observation may correspond to a push that contacts the target, misses it, or moves an occluded object into a recoverable pose, so success depends on object permanence and delayed contact reasoning. \textbf{Aliased Maze}: navigation through repeated corridor textures where topologically distinct locations produce nearly identical egocentric observations but require different actions, testing whether the belief state preserves topological counterfactuals. \textbf{Deferred Kitchen}: a long-horizon manipulation routine in which early interactions determine whether later subtasks remain feasible, often by changing hidden preconditions without immediate visual evidence. We evaluate normalized return and model-exploitation rate, defined as the percentage of failures where the planner selects a trajectory with high predicted value but low realized value.

\subsection{Baselines}

We compare against five latent world-model baselines. \textbf{RSSM-Recon}: a recurrent state-space model trained primarily with observation reconstruction and latent consistency. \textbf{Reward-aware RSSM}: adds reward and value prediction, testing whether task supervision alone recovers intervention structure. \textbf{CPC-RSSM}: adds a generic contrastive predictive objective, separating the effect of contrastive learning from counterfactual negative construction. \textbf{Dreamer-style}: an imagination-based latent world model trained for actor-critic learning in latent space. \textbf{TD-MPC-style}: a decoder-free latent control model optimized for value-guided planning.

\subsection{Metrics}

We report three classes of metrics. \textbf{Task performance}: episode success rate on Occluded Push and Aliased Maze, normalized return on Deferred Kitchen. \textbf{Counterfactual separability} (CFS) measures whether latent rollouts separate intervention outcomes:
\begin{equation}
    \mathrm{CFS}=\mathrm{AUROC}\left(d_z(\Phi(h,\alpha),\Phi(h,\alpha')),\mathbb{1}[y(h,\alpha)\neq y(h,\alpha')]\right),
\end{equation}
where $\Phi(h,\alpha)$ is the imagined latent future induced by intervention sequence $\alpha$ from history $h$, and $y(h,\alpha)$ is the corresponding intervention outcome. CFS is computed on the imagined dynamics latents $\Phi$ with the projection head $g_\theta$ discarded, so it measures the geometry the planner actually uses rather than the embedding the contrastive loss directly shapes. The projection head is a training-time device only. Outcome labels $y$ at evaluation come from privileged simulator state, exactly as during training. \textbf{Model-exploitation rate} measures planning failures caused by overoptimistic imagined trajectories: a model can produce high predicted values while remaining unsafe for planning if it collapses distinct action-conditioned futures.

\section{Results}
\label{sec:results}

\subsection{CLWM improves planning in the regime where prediction is ambiguous}

Table~\ref{tab:main_results} shows the main results. \clwm\ achieves the best performance on every benchmark and every diagnostic, gaining $9.5$ points on Occluded Push and $11.6$ on Aliased Maze over the strongest non-counterfactual baseline, raising Deferred Kitchen return from $62.8$ to $72.3$, and cutting model exploitation from $18.4\%$ to $9.7\%$. The pattern is consistent with the paper's hypothesis: the largest improvements occur in tasks where visually plausible prediction is not enough to support correct action selection.

\begin{table}[t]
\centering
\caption{Main results on partially observable embodied benchmarks. Success denotes episode success rate. CFS is counterfactual separability AUROC. Exploit rate is the percentage of planning failures caused by high predicted value but low realized value. All success and return values are means over 5 random seeds (seeds 0--4), and $\pm$ denotes the standard error of the mean. CFS and exploitation rate are computed over the same seeds. All methods are evaluated from their final checkpoints on 100 fixed test episodes. Aliased Maze results use aliasing intensity $0.0$, the benchmark's default layout (details in Appendix~\ref{app:details}).}
\label{tab:main_results}
\small
\setlength{\tabcolsep}{4pt}
\begin{tabular}{lcccccc}
\toprule
\multirow{2}{*}{Method} & \multicolumn{2}{c}{Occluded Push} & \multicolumn{2}{c}{Aliased Maze} & \multicolumn{2}{c}{Deferred Kitchen} \\
\cmidrule(lr){2-3}\cmidrule(lr){4-5}\cmidrule(lr){6-7}
& Success $\uparrow$ & CFS $\uparrow$ & Success $\uparrow$ & CFS $\uparrow$ & Return $\uparrow$ & Exploit $\downarrow$ \\
\midrule
RSSM-Recon & 47.8 $\pm$ 2.1 & 0.61 & 52.4 $\pm$ 1.8 & 0.58 & 43.2 $\pm$ 2.7 & 31.5 \\
Reward-aware RSSM & 55.6 $\pm$ 1.9 & 0.66 & 58.1 $\pm$ 2.0 & 0.63 & 51.7 $\pm$ 2.4 & 24.8 \\
CPC-RSSM & 58.4 $\pm$ 2.4 & 0.70 & 61.5 $\pm$ 1.7 & 0.68 & 53.9 $\pm$ 2.1 & 22.7 \\
Dreamer-style & 62.9 $\pm$ 1.6 & 0.69 & 65.7 $\pm$ 1.9 & 0.67 & 59.4 $\pm$ 1.8 & 20.9 \\
TD-MPC-style & 65.1 $\pm$ 1.6 & 0.71 & 67.3 $\pm$ 1.5 & 0.70 & 62.8 $\pm$ 2.0 & 18.4 \\
\midrule
\clwm\ (ours) & \textbf{74.6 $\pm$ 1.5} & \textbf{0.84} & \textbf{78.9 $\pm$ 1.4} & \textbf{0.86} & \textbf{72.3 $\pm$ 1.6} & \textbf{9.7} \\
\bottomrule
\end{tabular}
\end{table}

Two comparisons are especially informative. Reward-aware RSSM improves over RSSM-Recon but remains far below \clwm: reward supervision is sparse and trajectory-local, and does not by itself separate alternative futures before reward is observed. CPC-RSSM improves CFS relative to RSSM-Recon but still underperforms \clwm: generic contrastive prediction helps representation learning, but does not target the hard cases that matter for embodied control---same-belief interventions and perceptual aliases with different downstream outcomes. The exploitation reduction is also central: TD-MPC-style planning achieves the strongest baseline return on Deferred Kitchen yet still has an exploitation rate of $18.4\%$ among failed episodes. \clwm\ nearly halves this rate, changing not just average performance but the failure modes the planner encounters.

\subsection{Counterfactual separability tracks downstream control}

\begin{figure}[t]
\centering
\begin{tikzpicture}
\begin{axis}[
    width=0.72\linewidth,
    height=0.38\linewidth,
    xlabel={Counterfactual separability (AUROC)},
    ylabel={Planning success (\%)},
    xmin=0.55,xmax=0.90,
    ymin=40,ymax=85,
    grid=both,
    legend style={at={(0.03,0.97)},anchor=north west,draw=none,fill=none},
]
\addplot+[only marks,mark=*] coordinates {(0.61,47.8) (0.66,55.6) (0.70,58.4) (0.69,62.9) (0.71,65.1) (0.84,74.6)};
\addlegendentry{Occluded Push}
\addplot+[only marks,mark=square*] coordinates {(0.58,52.4) (0.63,58.1) (0.68,61.5) (0.67,65.7) (0.70,67.3) (0.86,78.9)};
\addlegendentry{Aliased Maze}
\addplot[domain=0.56:0.88,samples=2,dashed,thick,black!60] {115.9*x-17.3};
\addlegendentry{trend (baselines)}
\end{axis}
\end{tikzpicture}
\caption{Counterfactual separability tracks downstream planning success. Across the five baseline model classes---none of which optimizes CFS---the correlation between CFS and success is $r=0.94$ (Occluded Push) and $r=0.95$ (Aliased Maze). The dashed line is a least-squares fit to the ten baseline points. \clwm\ (rightmost point in each series) directly optimizes counterfactual structure, so its points illustrate consistency with the trend rather than independent validation of the metric.}
\label{fig:cfs_success}
\end{figure}
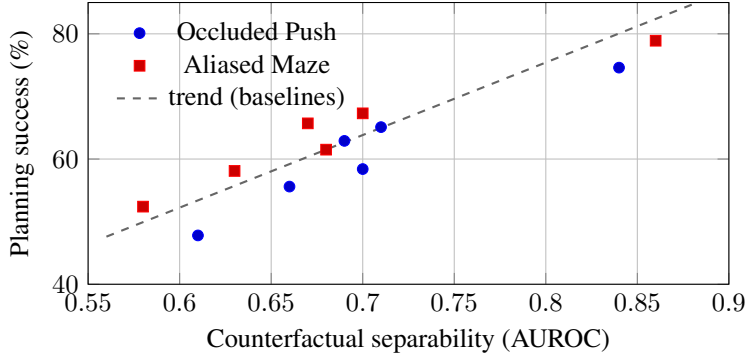

The main results reveal a strong alignment between CFS and planning performance: methods with low CFS also exhibit lower success and higher exploitation. Two caveats govern how much this alignment can show. Because \clwm\ directly optimizes an objective aligned with CFS, its own point in Figure~\ref{fig:cfs_success} cannot serve as independent validation of the metric. The informative evidence is the trend across the five baselines, none of which is trained on CFS. Across those baselines, the correlation between CFS and planning success is $r=0.94$ on Occluded Push and $r=0.95$ on Aliased Maze. This relationship is consistent with CFS capturing a property that reconstruction loss and return alone can obscure: whether imagined latent distance reflects meaningful differences in intervention outcome. We do not claim that CFS alone characterizes all aspects of world-model quality.

\subsection{Ablations show that the gains come from counterfactual structure}

Table~\ref{tab:ablations} isolates the contribution of each component of the counterfactual objective. The bottom row repeats the strongest non-counterfactual baseline from Table~\ref{tab:main_results} as a reference. The gap between full \clwm\ and this reference ($9.5$ points on Occluded Push, $11.6$ on Aliased Maze) is the quantity the ablations decompose. The most revealing ablation removes perceptual-alias negatives: success drops to $67.5\%$ and $69.1\%$, with average CFS falling from $0.85$ to $0.74$---precisely the failure mode the method is designed to address. When the model no longer receives pressure to distinguish futures that look similar but behave differently, it again collapses the cases that matter most for planning.

\begin{table}[t]
\centering
\caption{Ablation study. All ablation variants are trained and evaluated identically on all three benchmarks. Each column reports one benchmark's metric: success on Occluded Push, success on Aliased Maze, CFS averaged over those two tasks (rounded to two decimals, with per-task values for \clwm\ and the reference in Table~\ref{tab:main_results}), and exploitation rate on Deferred Kitchen. The bottom row is not an ablation: it repeats the strongest non-counterfactual baseline from Table~\ref{tab:main_results} for reference.}
\label{tab:ablations}
\small
\setlength{\tabcolsep}{4.2pt}
\renewcommand{\arraystretch}{1.02}
\begin{tabular}{lcccc}
\toprule
\textbf{Variant} &
\textbf{Push Succ.} $\uparrow$ &
\textbf{Maze Succ.} $\uparrow$ &
\textbf{Avg.\ CFS} $\uparrow$ &
\textbf{Kitchen Exploit} $\downarrow$ \\
\midrule
Full \clwm & \textbf{74.6} & \textbf{78.9} & \textbf{0.85} & \textbf{9.7} \\
No same-belief negatives & 70.2 & 72.4 & 0.79 & 13.6 \\
No perceptual-alias negatives & 67.5 & 69.1 & 0.74 & 16.8 \\
No value-disagreement negatives & 71.3 & 74.0 & 0.81 & 12.9 \\
No outcome head & 70.8 & 73.5 & 0.80 & 13.2 \\
Random negatives only & 68.1 & 70.6 & 0.76 & 15.4 \\
\midrule
Reference: TD-MPC-style (Table~\ref{tab:main_results}) & 65.1 & 67.3 & 0.71 & 18.4 \\
\bottomrule
\end{tabular}
\vspace{-0.5em}
\end{table}

Same-belief and value-disagreement negatives contribute complementary benefits: removing the former weakens separation of alternative actions from the same inferred state, while removing the latter raises exploitation to $12.9\%$. Random negatives are insufficient---they improve over the reference but remain far below full \clwm---showing that negative construction quality is central to the method.

Removing the outcome head ($\Linv$) costs $3.8$ and $5.4$ points while remaining well above the reference. Privileged outcome labels enter \clwm\ both through $\Linv$ and through the construction of contrastive pairs. The small cost of removing $\Linv$ indicates that their value is realized mainly through the contrastive geometry rather than through direct outcome prediction. We caution that Table~\ref{tab:ablations} does not include a variant that retains the outcome head without the contrastive loss ($\lambda_{\mathrm{cf}}=0$), nor a baseline given the same labels (e.g., TD-MPC-style with an outcome head). These supervision-matched controls are the right way to fully separate label access from representation geometry, and we leave them for the extended version.

\subsection{Robustness under perceptual aliasing}
\label{sec:aliasing_robustness}

We next stress-test the models by increasing perceptual aliasing intensity in Aliased Maze: as aliasing increases, local observations become less informative about the agent's true topological state, directly testing whether the model relies on surface-level observation prediction or preserves history-dependent intervention structure.

\begin{figure}[t]
\centering
\begin{tikzpicture}
\begin{axis}[
    width=0.76\linewidth,
    height=0.38\linewidth,
    xlabel={Perceptual aliasing intensity},
    ylabel={Success rate (\%)},
    xmin=0,xmax=1,
    ymin=30,ymax=92,
    grid=both,
    legend columns=2,
    legend style={at={(0.97,0.97)},anchor=north east,draw=none,fill=none,/tikz/every even column/.append style={column sep=0.4em}},
]
\addplot+[mark=*] coordinates {(0.0,52.4) (0.25,48.3) (0.50,42.0) (0.75,39.5) (1.0,38.0)};
\addlegendentry{RSSM-Recon}
\addplot+[mark=square*] coordinates {(0.0,65.7) (0.25,61.1) (0.50,55.5) (0.75,51.8) (1.0,50.0)};
\addlegendentry{Dreamer-style}
\addplot+[mark=triangle*] coordinates {(0.0,67.3) (0.25,63.5) (0.50,59.0) (0.75,56.2) (1.0,55.0)};
\addlegendentry{TD-MPC-style}
\addplot+[mark=diamond*] coordinates {(0.0,78.9) (0.25,75.2) (0.50,70.8) (0.75,68.1) (1.0,66.0)};
\addlegendentry{\clwm}
\end{axis}
\end{tikzpicture}
\caption{Robustness to perceptual aliasing. All results use the same final checkpoints and evaluation protocol as Table~\ref{tab:main_results} (100 test episodes, default maze layout). Intensity $0.0$ corresponds to the Table~\ref{tab:main_results} setting. Points are means over 5 seeds. All methods degrade as local observations become less informative. \clwm\ remains the strongest at every intensity and retains the largest fraction of its default-setting performance, consistent with latent dynamics trained to preserve action-conditioned distinctions that are not visible in the current observation.}
\label{fig:aliasing}
\end{figure}
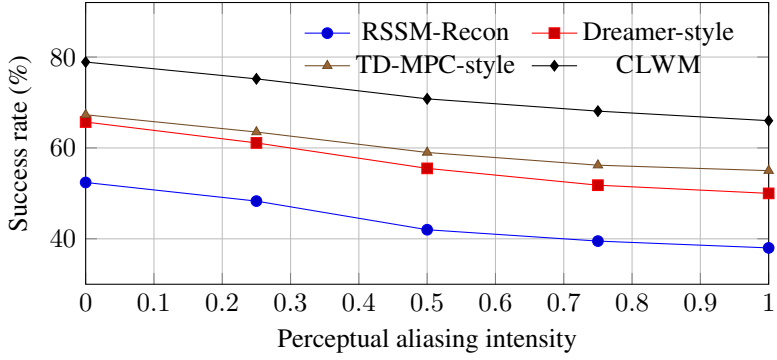

Figure~\ref{fig:aliasing} shows that all methods degrade as aliasing increases, but \clwm\ remains the strongest at every intensity: at the highest aliasing level it maintains $66\%$ success, compared with $55\%$ for TD-MPC-style planning and $38\%$ for RSSM-Recon. Its margin over the strongest baseline is essentially undiminished at maximum aliasing ($11.6$ points at intensity $0.0$, $11.0$ at intensity $1.0$), and it retains the largest fraction of its default-setting performance ($84\%$, versus $82\%$, $76\%$, and $73\%$ for TD-MPC-style, Dreamer-style, and RSSM-Recon). The gain is therefore not an artifact of the easy regime: it persists as local observations become uninformative and prediction alone is least sufficient.

\section{Discussion}
\label{sec:discussion}

The results support the paper's central claim: embodied world models should be evaluated by whether they preserve the consequences of possible actions, not only by whether they predict plausible observations. The gains arise in settings where these criteria diverge---occlusion, perceptual aliasing, hidden contact, and delayed feasibility---and persist essentially undiminished as aliasing intensifies. Neither task supervision nor generic contrastive learning suffices on its own, and the outcome-head ablation suggests the privileged labels' value is realized through the contrastive geometry rather than direct outcome prediction---though fully supervision-matched controls remain future work. The exploitation results clarify why this matters for control: if a latent model collapses two futures that differ in hidden feasibility, a planner can assign high value to an invalid action sequence. By reducing exploitative failures from $18.4\%$ to $9.7\%$, \clwm\ changes not only average performance but the way the model fails. We do not claim that \clwm\ learns a structural causal model or guarantees identifiability. The claim is practical: a world model for embodied planning should be trained to preserve the intervention distinctions its planner needs.

Finally, we note an implication for agents built on pretrained representations. Our experiments train latent dynamics from scratch, but nothing in the definition of counterfactual collapse depends on that choice. The CFS metric requires only a way to embed action-conditioned futures and a source of intervention-outcome labels. It can be computed for a frozen pretrained encoder, a fine-tuned one, or a policy's internal features. Given that Figure~\ref{fig:cfs_success} shows CFS tracking planning success across baseline model classes, we suggest measuring it before a pretrained representation is handed to a planner, rather than inferring actionability from downstream reward alone: reward is a slow and confounded signal for this property, and a representation can produce adequate average return while remaining exploitable in precisely the ambiguous states where an agent most needs a reliable model.

\section{Limitations and broader impact}
\label{sec:limitations}

\clwm\ requires reliable intervention-outcome signals: available from privileged state in simulation, but requiring trackers, tactile events, learned predicates, or human labels in the real world. It also increases training cost by sampling multiple interventions per belief state and mining hard negatives, though this can be reduced by reusing planner candidates or applying the objective only in ambiguous states. \clwm\ does not solve long-horizon compounding error, exploration, distribution shift, or reward misspecification, and should be viewed as complementary to uncertainty-aware model-based RL. Our claim about pretrained representations is conceptual rather than empirical: we characterize why counterfactual collapse should be expected under standard pretraining objectives and show that CFS is computable for any representation, but we do not measure CFS on large-scale pretrained encoders here. We regard that as the natural next experiment. Configuration details are collected in Appendix~\ref{app:details}.

World models that better preserve action consequences could improve embodied agents in assistive robotics, manipulation, and navigation, and may reduce overconfident planning through visually plausible but physically invalid futures. At the same time, more capable embodied planning increases the autonomy of physical systems, which requires conservative deployment practices: uncertainty estimation, human override, environment-specific validation, and clear limits on autonomous authority.

\section{Conclusion}
\label{sec:conclusion}

World models are usually trained to predict what is likely to happen next. Embodied agents need something stronger: models that preserve what would happen if they acted differently. \clwm\ combines recurrent belief inference, action-conditioned latent dynamics, and contrastive supervision over intervention outcomes to target counterfactual collapse---the failure to distinguish behaviorally different futures that look similar in observation space. Empirically, \clwm\ improves planning on all three partially observable benchmarks (by up to $11.6$ success points), increases counterfactual separability, and nearly halves model-exploitation failures, with ablations attributing the gains to hard perceptual-alias negatives and the advantage persisting essentially undiminished under maximum perceptual aliasing.

These gains also come at no deployment cost. The counterfactual machinery---intervention sampling, hard-negative mining, the projection head---exists only at training time, so the planner pays nothing at inference for the improved latent geometry. The objective is likewise orthogonal to the control loop: it attaches to any latent world model that rolls out action-conditioned futures, which makes it a drop-in training-time addition to existing world-model stacks rather than a competing architecture.

Looking forward, the most portable piece of this work may be the diagnostic rather than the training objective. CFS requires only imagined action-conditioned futures and intervention-outcome labels, so it can audit any inherited representation---frozen, fine-tuned, or pretrained at scale---before a planner is asked to trust it, and our results provide an initial calibration for that audit: across world-model classes trained from scratch, counterfactual separability tracked planning success at $r\geq 0.94$. A world model useful for planning must preserve not just what is likely to happen, but the intervention consequences of possible actions---and that property can, and should, be measured before it is assumed.

\clearpage
\begin{ack}
This work received no external funding. The authors declare no competing interests.
\end{ack}

\bibliographystyle{plainnat}

\begin{thebibliography}{99}

\bibitem[Ha and Schmidhuber(2018)]{ha2018worldmodels}
David Ha and Jürgen Schmidhuber.
\newblock Recurrent world models facilitate policy evolution.
\newblock In \emph{Advances in Neural Information Processing Systems}, 2018.

\bibitem[Hafner et~al.(2019)]{hafner2019planet}
Danijar Hafner, Timothy Lillicrap, Ian Fischer, Ruben Villegas, David Ha, Honglak Lee, and James Davidson.
\newblock Learning latent dynamics for planning from pixels.
\newblock In \emph{International Conference on Machine Learning}, 2019.

\bibitem[Hafner et~al.(2020)]{hafner2020dreamer}
Danijar Hafner, Timothy Lillicrap, Jimmy Ba, and Mohammad Norouzi.
\newblock Dream to control: Learning behaviors by latent imagination.
\newblock In \emph{International Conference on Learning Representations}, 2020.

\bibitem[Hafner et~al.(2023)]{hafner2023dreamerv3}
Danijar Hafner, Jurgis Pasukonis, Jimmy Ba, and Timothy Lillicrap.
\newblock Mastering diverse domains through world models.
\newblock arXiv preprint arXiv:2301.04104, 2023.

\bibitem[Hansen et~al.(2024)]{hansen2024tdmpc2}
Nicklas Hansen, Hao Su, and Xiaolong Wang.
\newblock TD-MPC2: Scalable, robust world models for continuous control.
\newblock In \emph{International Conference on Learning Representations}, 2024.

\bibitem[Wu et~al.(2022)]{wu2022daydreamer}
Philipp Wu, Alejandro Escontrela, Danijar Hafner, Ken Goldberg, and Pieter Abbeel.
\newblock DayDreamer: World models for physical robot learning.
\newblock arXiv preprint arXiv:2206.14176, 2022.

\bibitem[Grimm et~al.(2020)]{grimm2020value}
Christopher Grimm, Andr\'e Barreto, Satinder Singh, and David Silver.
\newblock The value equivalence principle for model-based reinforcement learning.
\newblock In \emph{Advances in Neural Information Processing Systems}, 2020.

\bibitem[Schrittwieser et~al.(2020)]{schrittwieser2020muzero}
Julian Schrittwieser, Ioannis Antonoglou, Thomas Hubert, Karen Simonyan, Laurent Sifre, Simon Schmitt, Arthur Guez, Edward Lockhart, Demis Hassabis, Thore Graepel, Timothy Lillicrap, and David Silver.
\newblock Mastering Atari, Go, chess and shogi by planning with a learned model.
\newblock \emph{Nature}, 588(7839):604--609, 2020.

\bibitem[Ferns et~al.(2004)]{ferns2004metrics}
Norm Ferns, Prakash Panangaden, and Doina Precup.
\newblock Metrics for finite Markov decision processes.
\newblock In \emph{Uncertainty in Artificial Intelligence}, 2004.

\bibitem[Gelada et~al.(2019)]{gelada2019deepmdp}
Carles Gelada, Saurabh Kumar, Jacob Buckman, Ofir Nachum, and Marc~G. Bellemare.
\newblock DeepMDP: Learning continuous latent space models for representation learning.
\newblock In \emph{International Conference on Machine Learning}, 2019.

\bibitem[Zhang et~al.(2021)]{zhang2021invariant}
Amy Zhang, Rowan McAllister, Roberto Calandra, Yarin Gal, and Sergey Levine.
\newblock Learning invariant representations for reinforcement learning without reconstruction.
\newblock In \emph{International Conference on Learning Representations}, 2021.

\bibitem[Khosla et~al.(2020)]{khosla2020supcon}
Prannay Khosla, Piotr Teterwak, Chen Wang, Aaron Sarna, Yonglong Tian, Phillip Isola, Aaron Maschinot, Ce Liu, and Dilip Krishnan.
\newblock Supervised contrastive learning.
\newblock In \emph{Advances in Neural Information Processing Systems}, 2020.

\bibitem[Janner et~al.(2019)]{janner2019mbpo}
Michael Janner, Justin Fu, Marvin Zhang, and Sergey Levine.
\newblock When to trust your model: Model-based policy optimization.
\newblock In \emph{Advances in Neural Information Processing Systems}, 2019.

\bibitem[Pearl(2009)]{pearl2009causality}
Judea Pearl.
\newblock \emph{Causality: Models, Reasoning, and Inference}.
\newblock Cambridge University Press, 2nd edition, 2009.

\bibitem[Peters et~al.(2017)]{peters2017elements}
Jonas Peters, Dominik Janzing, and Bernhard Schölkopf.
\newblock \emph{Elements of Causal Inference: Foundations and Learning Algorithms}.
\newblock MIT Press, 2017.

\bibitem[Li et~al.(2020)]{li2020causalworldmodels}
Minne Li, Mengyue Yang, Furui Liu, Xu Chen, Zhitang Chen, and Jun Wang.
\newblock Causal world models by unsupervised deconfounding of physical dynamics.
\newblock arXiv preprint arXiv:2012.14228, 2020.

\bibitem[Dasgupta et~al.(2019)]{dasgupta2019causalrl}
Ishita Dasgupta, Jane Wang, Silvia Chiappa, Jovana Mitrovic, Pedro Ortega, David Raposo, Edward Hughes, Peter Battaglia, Matthew Botvinick, and Zeb Kurth-Nelson.
\newblock Causal reasoning from meta-reinforcement learning.
\newblock arXiv preprint arXiv:1901.08162, 2019.

\bibitem[van den Oord et~al.(2018)]{oord2018cpc}
Aaron van den Oord, Yazhe Li, and Oriol Vinyals.
\newblock Representation learning with contrastive predictive coding.
\newblock arXiv preprint arXiv:1807.03748, 2018.

\bibitem[Laskin et~al.(2020)]{laskin2020curl}
Michael Laskin, Aravind Srinivas, and Pieter Abbeel.
\newblock CURL: Contrastive unsupervised representations for reinforcement learning.
\newblock In \emph{International Conference on Machine Learning}, 2020.

\bibitem[Kaelbling et~al.(1998)]{kaelbling1998pomdp}
Leslie Pack Kaelbling, Michael L. Littman, and Anthony R. Cassandra.
\newblock Planning and acting in partially observable stochastic domains.
\newblock \emph{Artificial Intelligence}, 101(1--2):99--134, 1998.

\end{thebibliography}

\appendix

\section{Experimental details}
\label{app:details}

Tables~\ref{tab:config} and~\ref{tab:impl} list the configuration and implementation details for all reported runs.

\begin{table}[h]
\centering
\caption{Training and evaluation configuration.}
\label{tab:config}
\small
\begin{tabularx}{\linewidth}{lX}
\toprule
Setting & Value \\
\midrule
Seeds \& statistics & 5 seeds (0--4); $\pm$ is the standard error of the mean over seeds \\
Belief / latent dimension & 200 / 32 \\
Planner & MPC with $N=512$ candidate sequences, horizon $H=12$ \\
Counterfactual rollout horizon & $K=8$ \\
Contrastive temperature & $\tau=0.07$ \\
Loss weights & $\lambda_{\mathrm{cf}}=0.1$, $\lambda_{\mathrm{inv}}=0.01$, $\beta_{\mathrm{KL}}=0.5$ (with free bits), $\beta_r=1.0$, $\beta_v=0.1$ \\
Optimization & AdamW, learning rate $3\times 10^{-4}$ with cosine decay to $10^{-6}$ and 10k-step warmup; batch size 256; $10^{6}$ environment steps \\
Evaluation & final checkpoints on 100 fixed test episodes (standard test split); Aliased Maze at default aliasing intensity $0.0$ \\
\bottomrule
\end{tabularx}
\end{table}

\begin{table}[h]
\centering
\caption{Implementation details.}
\label{tab:impl}
\small
\begin{tabularx}{\linewidth}{lX}
\toprule
Field & Value \\
\midrule
Counterfactual branching & $M=64$ intervention sequences per anchor; branch source: 50\% policy rollouts, 25\% random shooting, 25\% replay trajectories \\
Negative mining & perceptual-alias similarity via LPIPS on image patches over a 5-step window; 16 negatives per anchor; outcome distance threshold $\epsilon_y=0.1$ \\
Outcome descriptors $y$ & Occluded Push: hidden-object $(x,y)$ pose relative to goal, success threshold 0.05\,m; Aliased Maze: discrete topological cell index (0--63); Deferred Kitchen: binary feasibility vector over later subtasks (e.g., stove on, kettle filled) \\
Simulators & Occluded Push: MuJoCo (MJX) v1.2.1; Aliased Maze: Gymnasium-based grid world v0.1.0; Deferred Kitchen: RoboSuite v2.0 (adapted) \\
Encoder $e_\theta$ & 4-layer CNN (32/64/128/256 channels, kernel 3, stride 2, ReLU) followed by a fully connected layer to 256 units; proprioceptive/tactile branch: 2-layer MLP (128, 64), concatenated \\
Decoder & transposed CNN mirroring the encoder \\
Projection head $g_\theta$ & 2-layer MLP applied to the 256-dimensional rollout summary $\Phi$: $256 \to 128 \to 64$ (ReLU); used only in the contrastive loss and discarded at evaluation \\
Outcome head & 2-layer MLP: $256 \to 128 \to$ number of outcome classes (ReLU; softmax applied in the loss) \\
Sensitivity & $\lambda_{\mathrm{cf}}\in[0.05,0.2]$ and $\tau\in[0.05,0.1]$ swept; chosen values ($0.1$, $0.07$) stable across the range \\
Compute & single NVIDIA A100; $10^{6}$ environment steps; $\sim$24\,h per \clwm\ run, $\sim$12\,h per baseline run \\
\bottomrule
\end{tabularx}
\end{table}

\end{document}